\documentclass{article}
\usepackage{spconf,amsmath,graphicx}
\usepackage[margin=1in,bottom=1.10in,footskip=0.6in,columnsep=0.32in]{geometry} 
\usepackage{enumitem}
\usepackage{subcaption}   
\setlist{nosep, leftmargin=14pt}

\usepackage{mwe} 
\makeatletter
\renewcommand\section{\@startsection {section}{1}{\z@}%
  {2.5ex plus .8ex minus .4ex}
  {1.5ex plus .5ex minus .3ex}
  {\normalfont\Large\bfseries}}

\renewcommand\subsection{\@startsection {subsection}{2}{\z@}%
  {2.0ex plus .6ex minus .3ex}
  {1.0ex plus .3ex minus .2ex}
  {\normalfont\large\bfseries}}

\renewcommand\subsubsection{\@startsection {subsubsection}{3}{\z@}%
  {1.5ex plus .4ex minus .2ex}
  {0.8ex plus .2ex minus .1ex}
  {\normalfont\normalsize\bfseries}}
\makeatother

\title{Spatially-Aware Mixture of Experts with Log-Logistic Survival Modeling for Whole-Slide Images}
\name{
\begin{tabular}{c}
    \textit{Ardhendu Sekhar}, \textit{Vasu Soni}, \textit{Keshav Aske}, \textit{Shivam Madnoorkar}, \textit{Pranav Jeevan}, \textit{Amit Sethi}
\end{tabular}
}
\address{Indian Institute of Technology Bombay}
\begin{document}
%
\maketitle

\begin{abstract}
Accurate survival prediction from histopathology whole-slide images (WSIs) remains challenging due to gigapixel resolutions, spatial heterogeneity, and complex survival distributions. We introduce a comprehensive computational pathology framework that addresses these limitations through four synergistic innovations: (1) \textbf{Quantile-Gated Patch Selection} to dynamically identify prognostically relevant regions; (2) \textbf{Graph-Guided Clustering} that groups patches by spatial-morphological similarity to capture phenotypic diversity; (3) \textbf{Hierarchical Context Attention} to model both local tissue interactions and global slide-level context; and (4) an \textbf{Expert-Driven Mixture of Log-Logistics} module that flexibly models complex survival distributions. On large-scale TCGA cohorts, our method achieves state-of-the-art performance, with time-dependent concordance indices of $0.644{\pm}0.059$ on LUAD, $0.751{\pm}0.037$ on KIRC, and $0.752{\pm}0.011$ on BRCA—significantly outperforming both histology-only and multimodal benchmarks. The framework provides improved calibration and interpretability, advancing the potential of WSIs for personalized cancer prognosis.

\end{abstract}
\section{Introduction}
\label{sec:intro}

Accurate survival prediction from histopathology images is critical for personalized cancer care, guiding treatment selection and risk stratification. While clinical and genomic markers are routinely used, histopathology images contain rich, underutilized morphological information that could significantly enhance prognostic accuracy. Whole-slide images (WSIs) capture comprehensive tissue-level morphology—including tumor architecture, stromal composition, and microenvironmental context—but their gigapixel resolutions and lack of localized annotations make direct modeling computationally prohibitive. Patch-based multiple instance learning (MIL) has thus emerged as the dominant paradigm for weakly supervised analysis of WSIs.

However, survival prediction introduces unique challenges beyond conventional classification tasks. It requires modeling long-range spatial dependencies and subtle morphological patterns across distributed tissue regions, where the arrangement and interaction of different tissue phenotypes may carry critical prognostic significance. Conventional statistical approaches, such as Cox proportional hazards, struggle to capture the complex, high-dimensional, non-linear relationships in histopathology data, while deep learning methods often lack interpretability and fail to maintain spatial coherence across the slide.

Current MIL approaches for survival analysis face several limitations: they often treat patches as independent instances, ignoring spatial relationships; they lack mechanisms to disentangle heterogeneous tissue phenotypes; and they typically rely on simplistic survival modeling that cannot capture complex, multi-modal hazard distributions. These limitations restrict their ability to fully leverage the rich morphological information embedded in WSIs.

To overcome these challenges, we introduce a unified computational pathology framework that integrates four key innovations: (1) \textbf{Quantile-Gated Patch Selection} to dynamically identify prognostically relevant regions while filtering noise; (2) \textbf{Graph-Guided Clustering} to group patches by spatial-morphological similarity, capturing phenotypic diversity; (3) \textbf{Hierarchical Context Attention} to model both fine-grained local interactions and global slide-level context; and (4) an \textbf{Expert-Driven Mixture of Log-Logistics} that flexibly models complex survival distributions through specialized components. Together, these elements enable interpretable and accurate survival prediction by jointly leveraging spatial structure, phenotypic abstraction, and probabilistic outcome modeling, providing a more comprehensive approach to WSI-based prognosis.

\section{Related Work}

Weakly supervised learning via multiple instance learning (MIL) has been central to WSI-based survival modeling, addressing the challenge of slide-level labels with patch-level features. Early attention-based methods, such as ABMIL~\cite{ilse2018attentionbaseddeepmultipleinstance}, CLAM~\cite{lu2020dataefficientweaklysupervised}, and DSMIL~\cite{li2021dualstreammultipleinstancelearning}, identified prognostic patches but overlooked spatial context and inter-patch relationships. Subsequent approaches incorporated structural information through graph-based methods (PatchGCN~\cite{chen2021slideimages2dpoint}) and hierarchical transformers (HGT~\cite{10.1007/978-3-031-43987-2_72}, TransMIL~\cite{shao2021transmiltransformerbasedcorrelated}, HIPT~\cite{chen2022scalingvisiontransformersgigapixel}), capturing spatial and multi-scale relations at higher computational cost.

Recent efforts have explored multimodal integration of histopathology with genomic (PathoGen-X~\cite{10981028}), methylation (CoC~\cite{ZhoHai_CoC_MICCAI2025}), and language data (HiLa~\cite{CuiJia_HiLa_MICCAI2025}). In parallel, probabilistic methods, such as SCMIL~\cite{Yang_2024} employ Gaussian distributions for survival estimation but lack flexibility for complex hazard shapes. However, these approaches typically lack explicit mechanisms for phenotype disentanglement and struggle to model heterogeneous compositions in cancer microenvironments.

Our method addresses these limitations through integrated phenotype-aware clustering and expert-driven survival modeling. We combine graph-guided clustering for spatial-morphological coherence, hierarchical attention for multi-scale context, and a mixture of log-logistic experts that adapt to diverse survival patterns. This unified approach enables interpretable phenotype disentanglement while flexibly modeling complex survival distributions beyond conventional parametric assumptions.

\begin{figure*}[t]
    \centering
    \includegraphics[width=0.90\textwidth]{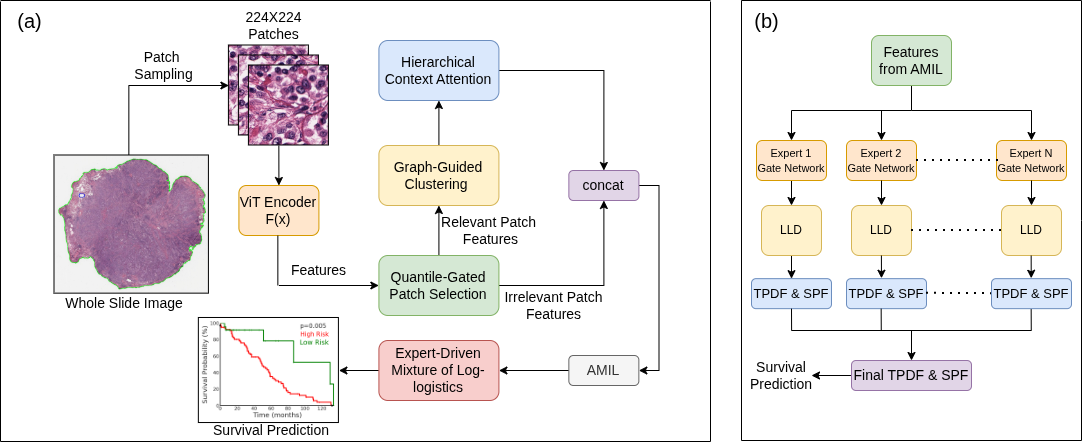}
    \caption{\textbf{(a)} Proposed survival framework. \textbf{(b)} Expert-Driven Mixture of Log-logistics block.}
    \label{fig:pipeline}
    \label{fig:pipeline}
\end{figure*}

\section{Methodology}\label{sec:method}

Figure~\ref{fig:pipeline} illustrates the overall methodology. Each Whole Slide Image (WSI) is divided into non-overlapping $224\times224$ patches at 40$\times$ magnification. Background regions are removed, and remaining patches are encoded using a histopathology foundation model~\cite{kang2023benchmarkingselfsupervisedlearningdiverse}, yielding a patch feature matrix $\mathbf{P}_{\text{feat}} \in \mathrm{R}^{n \times d}$ where $n$ is the number of patches and $d$ the feature dimensionality. 

\subsection{Quantile-Gated Patch Selection (QGPS)}
A quantile-based MLP selects task-relevant patches using importance scores from model logits, filtering out uninformative regions to reduce computational burden and noise. For each WSI, threshold $\tau_q$ is the $q$-quantile of logits, retaining the top $(1-q)\times 100\%$ of patches:
\begin{equation}
\mathcal{P}_{\text{sel}}=\{\mathbf{P}_i \mid \text{logit}_i>\tau_q\},\quad
\mathcal{P}_{\text{rem}}=\{\mathbf{P}_i \mid \text{logit}_i\le\tau_q\}.
\label{eq:sel}
\end{equation}

\subsection{Graph-Guided Clustering (GGC)}
Selected patches $\mathcal{P}_{\text{sel}}\in\mathrm{R}^{m\times d}$ are clustered into morphologically and spatially coherent groups to capture phenotypic diversity. Morphological similarity uses cosine similarity:
\begin{equation}
S_{\text{morph}}(i,j)=\frac{\langle \mathbf{P}_i, \mathbf{P}_j\rangle}{\|\mathbf{P}_i\|\,\|\mathbf{P}_j\|},
\label{eq:smorph}
\end{equation}
while spatial similarity uses an exponential kernel over Euclidean distances:
\begin{equation}
S_{\text{spatial}}(i,j)=\exp\!\left(-\frac{D_{ij}}{\sigma_D}\right),\quad \sigma_D=\text{std}(D)+\varepsilon.
\label{eq:sspatial}
\end{equation}
The composite similarity $S=\omega_{\text{morph}}S_{\text{morph}}+\omega_{\text{spatial}}S_{\text{spatial}}$ guides a $k$-NN graph construction, with patches clustered into $G$ groups ${L_1,\dots,L_G}$ using GPU-accelerated K-Means.

\subsection{Hierarchical Context Attention (HCA)}
Each cluster $L_i$ undergoes multi-head self-attention (MHSA)~\cite{vaswani2017attention} for local dependencies, capturing fine-grained relationships within phenotypically similar regions:
\vspace{-1.0em}
\begin{equation}
L'_i=\text{LayerNorm}\!\big(L_i+\text{MHSA}(L_i)\big).
\label{eq:intra}
\end{equation}
Refined clusters are summarized into embeddings $R_i=\frac{1}{|L'_i|}\sum_{\mathbf{x}\in L'_i}\mathbf{x}$, then processed by another MHSA layer for global relationships across different tissue regions:
\begin{equation}
R'=\text{LayerNorm}\!\big(R+\text{MHSA}(R)\big).
\label{eq:inter}
\end{equation}
Intra-cluster representations are concatenated as $\widetilde{P}=\mathrm{Concat}(L'_g)_{g=1}^{G}$. A broadcast descriptor $R'_{\mathrm{exp}}$, derived by averaging $R'$ across clusters, is added residually to form $\widehat{P}$, which is concatenated with remaining patches to yield $\mathcal{P}_{\text{final}}$.

Slide-level representation is derived via attention pooling:
\begin{equation}
\mathbf{z}_{\text{WSI}} = \sum_i \alpha_i\,\mathcal{P}_{\text{final},i},\quad 
\alpha_i = \text{softmax}\!\big(W_a\,\tanh(W_h\,\mathcal{P}_{\text{final},i}^{\top})\big),
\label{eq:wsiagg}
\end{equation}
where $\mathbf{z}_{\text{WSI}}$ represents the integrated prognostic feature for the entire slide.

\subsection{Expert-Driven Mixture of Log-logistics (EDMLL)}
The module models survival probabilities using multiple experts specializing in distinct WSI features $\mathbf{z}_{\text{WSI}}$, enabling capture of diverse risk patterns. Each expert models the conditional distribution using log-logistic mixtures:
\begin{equation}
p(t \mid \mathbf{z}_{\text{WSI}}, e) = \sum_{k=1}^{K} \lambda^{(e)}_k(\mathbf{z}_{\text{WSI}})\,\mathrm{LLD}\!\big(t \mid \alpha^{(e)}_k, \beta^{(e)}_k\big),
\label{eq:lld}
\end{equation}
where the log-logistic density is:
\begin{equation}
\mathrm{LLD}(t \mid \alpha, \beta) = \frac{(\beta/\alpha)(t/\alpha)^{\beta-1}}{\big[1 + (t/\alpha)^{\beta}\big]^2}, \quad t>0.
\label{eq:lld_pdf}
\end{equation}
Learnable global anchors $(P_\alpha, P_\beta) \in \mathrm{R}^K$ are projected as:
\begin{equation}
\alpha^{(e)} = \text{softplus}\!\big(W^{(e)}_\alpha P_\alpha\big), \quad
\beta^{(e)} = \text{softplus}\!\big(W^{(e)}_\beta P_\beta\big).
\label{eq:anchors_lld}
\end{equation}
A gating network $G(\mathbf{z}_{\text{WSI}})$ assigns softmax weights to experts, producing the time probability density(TPDF) and survival probability functions(SPF):
{\small
\begin{equation}
\text{TPDF}(t \mid \mathbf{z}_{\text{WSI}}) =
\sum_{e,k} G_e(\mathbf{z}_{\text{WSI}})\lambda^{(e)}_k\,
\mathrm{LLD}\!\big(t \mid \alpha^{(e)}_k,\beta^{(e)}_k\big),
\label{eq:tpdf_lld}
\end{equation}
\begin{equation}
\text{SPF}(t \mid \mathbf{z}_{\text{WSI}}) =
1-\!\sum_{e,k} G_e(\mathbf{z}_{\text{WSI}})\lambda^{(e)}_k
F_{\mathrm{LLD}}\!\big(t \mid \alpha^{(e)}_k,\beta^{(e)}_k\big).
\label{eq:spf_lld}
\end{equation}
}
where $F_{\mathrm{LLD}}$ is obtained by integrating the LLD.

The model is trained using negative log-likelihood loss handling both censored ($c=0$) and uncensored ($c=1$) data:
\begin{equation}
\begin{aligned}
\mathcal{L}_{\text{NLL}} = & -c\log\!\big(\text{TPDF}(t_d\mid \mathbf{z}_{\text{WSI}})\big) \\
                          & - (1-c)\log\!\big(\text{SPF}(t_d\mid \mathbf{z}_{\text{WSI}})\big),
\end{aligned}
\label{eq:nll}
\end{equation}
combined with gating-entropy loss: 
\begin{equation}
\mathcal{L}_{\text{ent}} = -\sum_e G_e(\mathbf{z}_{\text{WSI}})\log(G_e(\mathbf{z}_{\text{WSI}}))
\end{equation}
to encourage diverse expert usage. The total loss is $\mathcal{L}_{\text{total}}=\mathcal{L}_{\text{NLL}}+\lambda_{\text{ent}}\mathcal{L}_{\text{ent}}$.

\begin{figure*}[t]
    \centering
    \includegraphics[width=0.90\textwidth]{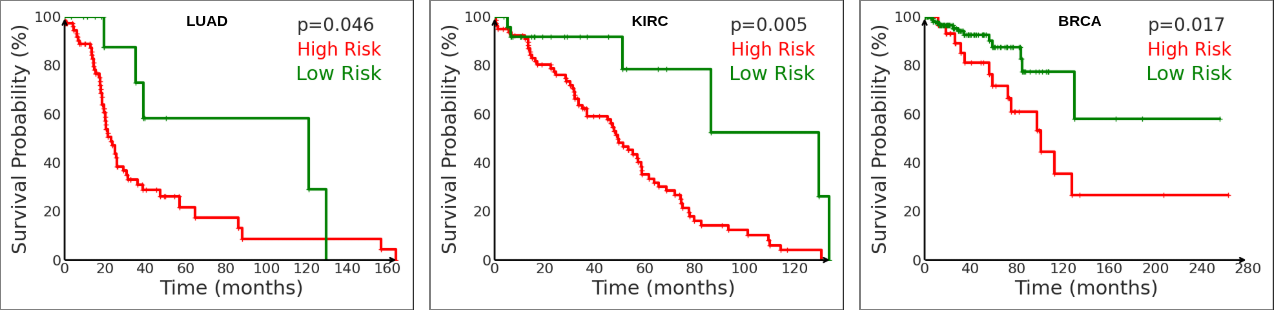}
    \caption{Kaplan–Meier survival curves showing significant high- vs. low-risk separation across TCGA cohorts.}
    \label{fig:km_combined}
\end{figure*}

\begin{table*}[t]
\centering
\captionsetup{font=small}
\caption{Left: TCGA performance (TDC, higher is better). Right: Ablations on $q$ in QGPS, GGC, HCA and $e$ in EDMLL. Best TDC results in \textbf{bold}.}
\label{tab:main+abl}
\begin{subtable}[t]{0.60\textwidth}
\centering
\scriptsize
\setlength{\tabcolsep}{5pt}
\renewcommand{\arraystretch}{1.05}
\caption{TCGA results. Modality:I=image,G=genomic,M=methylation,L=language prompt.}
\label{tab:results}
\begin{tabular}{l|c|c|c|c|c}
\hline
\textbf{Method} & \textbf{Modality} & \textbf{LUAD} & \textbf{KIRC} & \textbf{BRCA} & \textbf{MEAN} \\
\hline
AMIL~\cite{ilse2018attentionbaseddeepmultipleinstance} & I   & $0.632 \pm 0.021$ & $0.691 \pm 0.022$ & $0.735 \pm 0.013$ & $0.686$ \\
CLAM~\cite{lu2020dataefficientweaklysupervised}        & I   & $0.615 \pm 0.041$ & $0.680 \pm 0.028$ & $0.652 \pm 0.027$ & $0.649$ \\
DSMIL~\cite{li2021dualstreammultipleinstancelearning}  & I   & $0.609 \pm 0.039$ & $0.644 \pm 0.013$ & $0.668 \pm 0.011$ & $0.640$ \\
TransMIL~\cite{shao2021transmiltransformerbasedcorrelated} & I & $0.568 \pm 0.017$ & $0.642 \pm 0.076$ & $0.636 \pm 0.018$ & $0.615$ \\
PatchGCN~\cite{chen2021slideimages2dpoint}             & I   & $0.587 \pm 0.019$ & $0.675 \pm 0.045$ & $0.582 \pm 0.037$ & $0.615$ \\
HIPT~\cite{chen2022scalingvisiontransformersgigapixel} & I   & $0.549 \pm 0.025$ & $0.640 \pm 0.037$ & $0.625 \pm 0.046$ & $0.605$ \\
HGT~\cite{10.1007/978-3-031-43987-2_72}                & I   & $0.607 \pm 0.058$ & $0.648 \pm 0.018$ & $0.648 \pm 0.022$ & $0.634$ \\
SCMIL~\cite{Yang_2024}                                 & I   & $0.622 \pm 0.015$ & $0.688 \pm 0.037$ & $0.674 \pm 0.048$ & $0.661$ \\
OTSurv~\cite{RenQin_OTSurv_MICCAI2025}                                          & I   & $0.638 \pm 0.077$ & $0.750 \pm 0.149$ & $0.621 \pm 0.071$ & $0.670$ \\
\hline
PathoGen-X~\cite{10981028}                             & I+G & $0.620 \pm 0.008$ & $- - -$           & $0.670 \pm 0.020$ & $0.645$ \\
CoC~\cite{ZhoHai_CoC_MICCAI2025}                       & I+M & $- - -$           & $0.709 \pm 0.048$ & $0.654 \pm 0.036$ & $0.681$ \\
HiLa~\cite{CuiJia_HiLa_MICCAI2025}                     & I+L & $0.643 \pm 0.055$ & $- - -$           & $0.659 \pm 0.044$ & $0.651$ \\
\hline
\textbf{Ours}                                          & \textbf{I} & \textbf{0.644} $\pm$ \textbf{0.059} & \textbf{0.751} $\pm$ \textbf{0.037} & \textbf{0.752} $\pm$ \textbf{0.011} & \textbf{0.716} \\

\hline
\end{tabular}
\end{subtable}\hfill
\begin{subtable}[t]{0.37\textwidth}
\centering
\scriptsize
\setlength{\tabcolsep}{3.5pt}
\renewcommand{\arraystretch}{1.05}
\caption{Ablations.}
\label{tab:ablation}
\begin{tabular}{l|c|c|c}
\hline
\textbf{Variant} & \textbf{LUAD} & \textbf{KIRC} & \textbf{BRCA} \\
\hline
$q$=0.5 in QGPS  & 0.630$\pm$0.022 & 0.738$\pm$0.052 & 0.739$\pm$0.046 \\
$q$=0.75 in QGPS & 0.621$\pm$0.041 & 0.727$\pm$0.037 & 0.729$\pm$0.019 \\
w/o QGPS & 0.622$\pm$0.053 & 0.729$\pm$0.065 & 0.730$\pm$0.027 \\
w/o GGC, HCA  & 0.616$\pm$0.046 & 0.723$\pm$0.050 & 0.723$\pm$0.011 \\
$e$=1 in EDMLL & 0.628$\pm$0.017 & 0.726$\pm$0.033& 0.730$\pm$0.041\\
\hline
\textbf{Ours (All)} & \textbf{0.644}$\pm$\textbf{0.059} & \textbf{0.751}$\pm$\textbf{0.037} & \textbf{0.752}$\pm$\textbf{0.011} \\
\hline
\end{tabular}
\end{subtable}
\end{table*}

\section{Experiments and Results}
\label{sec:experiments}

We evaluate our framework on three TCGA cohorts: Lung Adenocarcinoma (LUAD; 459 WSIs), Kidney Renal Clear Cell Carcinoma (KIRC; 509 WSIs), and Breast Invasive Carcinoma (BRCA; 956 WSIs)~\cite{tcga2013pan}. Each dataset contains WSIs annotated with overall survival time, measured in months from diagnosis to death (uncensored) or last follow-up (censored). These cohorts represent diverse cancer types with distinct morphological characteristics, providing a comprehensive testbed for generalization.

The quantile hyperparameter is set to $q=0.25$, retaining the top $75\%$ of patches based on logits. The $k$-NN graph constructed from the similarity matrix $S$ connects each point to its 10 nearest neighbors. Based on the graph connectivity, the patch indices are clustered into $G$ groups of 64 each, and these grouped patches are subsequently processed by an 8-head MHSA within the HCA module. The Expert-Driven Mixture of Log-Logistics module employs five experts, each modeling a log-logistic distribution (LLD) with $K = 100$ components. The model is trained for 20 epochs with Adam (lr = $2\times10^{-4}$, weight decay = $1\times10^{-3}$, dropout = 0.1) and a batch size of 1. Results are reported as mean $\pm$ standard deviation of the Time-Dependent Concordance Index (TDC)~\cite{han2022survivalmixturedensitynetworks} metric over 5-fold cross-validation. All experiments were run on an NVIDIA A6000 GPU.
\vspace{-0.5em}
\subsection{Comparative Performance Analysis}
Table~\ref{tab:main+abl}a summarizes performance across TCGA datasets. Our framework consistently achieves the highest mean TDC, indicating superior discrimination and calibration. We compare against histology-based MIL models—AMIL~\cite{ilse2018attentionbaseddeepmultipleinstance}, CLAM~\cite{lu2020dataefficientweaklysupervised}, DSMIL~\cite{li2021dualstreammultipleinstancelearning}, and TransMIL~\cite{shao2021transmiltransformerbasedcorrelated}—which served as backbones for the expert-driven mixture of log-logistics block. Survival-specific MIL methods such as HIPT~\cite{chen2022scalingvisiontransformersgigapixel}, PatchGCN~\cite{chen2021slideimages2dpoint}, HGT~\cite{10.1007/978-3-031-43987-2_72}, SCMIL~\cite{Yang_2024} and OTSurv~\cite{RenQin_OTSurv_MICCAI2025} act as baselines. Since multimodal frameworks~\cite{10981028,ZhoHai_CoC_MICCAI2025,CuiJia_HiLa_MICCAI2025} lack public code, the results are taken from respective papers.

Our method attains TDC scores of $0.644\pm0.059$ on LUAD, $0.751\pm0.037$ on KIRC, and $0.752\pm0.011$ on BRCA, achieving state-of-the-art performance on all datasets. The consistent improvement across diverse cancer types demonstrates the robustness of our approach. By combining quantile-based patch selection, graph-guided clustering, hierarchical attention, and expert-driven mixture of log-logistics, our model effectively identifies survival-relevant regions. Remarkably, it surpasses multimodal systems while relying solely on histopathology images.
\vspace{-0.5em}
\subsection{Ablation Studies and Component Analysis}
Ablation studies evaluate the contributions of Quantile-Gated Patch Selection, Graph-Guided Clustering, Hierarchical Context Attention and Expert-Driven Mixture of Log-logistics (Table~\ref{tab:main+abl}b). Varying the quantile parameter shows that $q=0.25$ provides the best trade-off, preserving prognostic regions while filtering irrelevant tissue. Eliminating any module results in reduced TDC, confirming their complementary contributions to noise suppression and contextual feature integration. The single-expert configuration ($e=1$) underperforms compared to five experts, underscoring the benefit of multi-expert modeling.
\vspace{-0.5em}
\subsection{Clinical Validation and Interpretability}
For interpretability, patients are categorized into high- and low-risk groups based on predicted survival scores. The Kaplan-Meier curves (Fig.~\ref{fig:km_combined}) reveal distinct survival trends with statistically significant separation (log-rank $p=0.046$ for LUAD, $p=0.005$ for KIRC, $p=0.017$ for BRCA), confirming that the proposed method effectively captures prognostically relevant morphological patterns. The clear risk stratification demonstrates clinical utility for identifying patients who may benefit from more aggressive treatment.

\section{Conclusion}

We presented a unified computational pathology framework for WSI-based survival prediction that integrates quantile-based patch selection, graph-guided clustering, hierarchical context attention, and expert-driven mixture modeling. Our approach captures local–global tissue interactions through spatially coherent clustering and models complex survival distributions via specialized log-logistic experts, achieving both superior discrimination and calibration. Extensive validation on TCGA-LUAD, KIRC, and BRCA cohorts demonstrates consistent and statistically significant improvements, outperforming both pathology-based and multimodal benchmarks. Notably, our framework surpasses multimodal methods using histology alone, highlighting untapped prognostic value in morphological patterns. Future work will explore multimodal integration and uncertainty estimation for greater clinical robustness.



\section{Compliance with ethical standards}
\label{sec:ethics}

This study used publicly available human data from~\cite{tcga2013pan}; hence, no ethical approval was required under the associated open-access license.

\section{Acknowledgments}
The results of this study are based on the data collected from the public TCGA Research Network~\cite{tcga2013pan}.

\bibliographystyle{IEEEbib}
\bibliography{strings,refs}

\end{document}